\documentclass[10pt,twocolumn,letterpaper]{article}

\usepackage{iccv}
\usepackage{times}
\usepackage{epsfig}
\usepackage{graphicx}
\usepackage{amsmath}
\usepackage{amssymb}
\usepackage{setspace}
\usepackage{multirow}
\usepackage[accsupp]{axessibility}

\usepackage[pagebackref=true,breaklinks=true,letterpaper=true,bookmarks=false]{hyperref}

\iccvfinalcopy 

\def\iccvPaperID{3384} 

\ificcvfinal\fi

\begin{document}

\title{AGG-Net: Attention Guided Gated-convolutional Network for\\
Depth Image Completion}

\author{Dongyue Chen$^{1,2,3,*}$, Tingxuan Huang$^{1,*}$, Zhimin Song$^{1}$, Shizhuo Deng$^{1,2}$, Tong Jia$^{1,3}$\\
$^{1}$ College of Information Science and Engineering,\\
Northeastern University Shenyang 110819, Liaoning, China\\
$^{2}$ National Frontiers Science Center for Industrial Intelligence and Systems Optimization, \\
Northeastern University, Shenyang 110819, Liaoning, China\\
$^{3}$ Foshan Graduate School of Innovation,\\
Northeastern University, Foshan 528311, Guangdong, China\\
{\tt\small https://github.com/htx0601/AGG-Net}
}


\maketitle
\ificcvfinal
\thispagestyle{empty}\fi

\begin{abstract}
   Recently, stereo vision based on lightweight RGBD cameras has been widely used in various fields. However, limited by the imaging principles, the commonly used RGB-D cameras based on TOF, structured light, or binocular vision acquire some invalid data inevitably, such as weak reflection, boundary shadows, and artifacts, which may bring adverse impacts to the follow-up work. In this paper, we propose a new model for depth image completion based on the Attention Guided Gated-convolutional Network (AGG-Net), through which more accurate and reliable depth images can be obtained from the raw depth maps and the corresponding RGB images. Our model employs a UNet-like architecture which consists of two parallel branches of depth and color features. In the encoding stage, an Attention Guided Gated-Convolution (AG-GConv) module is proposed to realize the fusion of depth and color features at different scales, which can effectively reduce the negative impacts of invalid depth data on the reconstruction. In the decoding stage, an Attention Guided Skip Connection (AG-SC) module is presented to avoid introducing too many depth-irrelevant features to the reconstruction. The experimental results demonstrate that our method outperforms the state-of-the-art methods on the popular benchmarks NYU-Depth V2, DIML, and SUN RGB-D.\vspace{-0.1cm}
\end{abstract}

\section{Introduction}
\label{sec:intro}

\begin{figure}[t]
\centering
\rotatebox{90}{~~Raw depth}\vspace{2pt}
\includegraphics[width=0.3\linewidth]{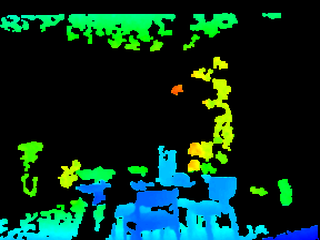}\vspace{3pt}
\includegraphics[width=0.3\linewidth]{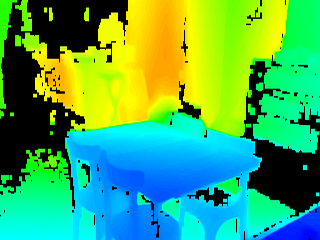}\vspace{3pt}
\includegraphics[width=0.3\linewidth]{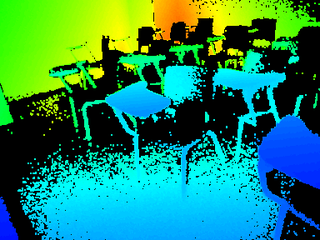}
\begin{spacing}{0.4}
\end{spacing}
\rotatebox{90}{Ours}\vspace{2pt}
\begin{minipage}{0.3\linewidth}
	\vspace{3pt}
	\centerline{\includegraphics[width=\textwidth]{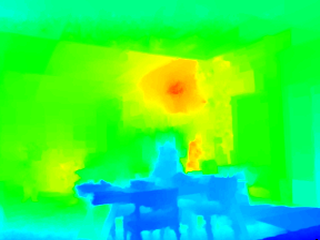}}
	\centerline{(a) NYU-Depth V2}
\end{minipage}
\begin{minipage}{0.3\linewidth}
	\vspace{3pt}
	\centerline{\includegraphics[width=\textwidth]{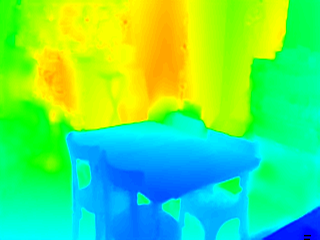}}
	\centerline{(b) DIML}
\end{minipage}
\begin{minipage}{0.3\linewidth}
	\vspace{3pt}
	\centerline{\includegraphics[width=\textwidth]{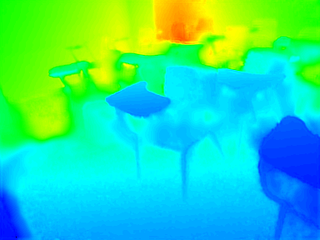}}
	\centerline{(c) SUN RGB-D}
\end{minipage}
\begin{spacing}{1.6}
\end{spacing}
\caption{Typical raw depth images with invalid data in the popular benchmark datasets (a) NYU-Depth V2, (b) DIML, and (c) SUN RGB-D, and the completion results of our method.}
\label{fig:fig1}
\end{figure}

Depth sensing is critical in applications such as autonomous driving \cite{Chen_2015_ICCV}, robot navigation \cite{1307213}, and scene reconstruction \cite{378185}. The commonly used depth sensors include LiDAR, Time-of-Flight, or binocular camera. However, most of the acquired depth images will inevitably be accompanied by many invalid areas caused by weak or direct reflections, far distance, bright light, and other environmental noises, as shown in Fig. \ref{fig:fig1}. These invalid data will have severe diverse impacts on the following process. Therefore, depth completion has become necessary for most applications based on depth images. 

Although many approaches based solely on raw depth images have been proposed, their performance is severely limited because of the absence or uncertainty of invalid data. Therefore, researchers consider introducing RGB information to guide the depth completion \cite{Ferstl_2013_ICCV} through two typical manners. The traditional one fills invalid pixels based on their valid neighbors according to some given rules, such as joint bilateral filters \cite{1467423}, fast marching \cite{doi:10.1137/S0036144598347059}, and Markov random field \cite{10.1007/BFb0028368}. However, these methods are generally not fast nor sufficiently accurate. The other approach predicts invalid pixels with the deep neural network, which usually employs an auto-encoder to extract depth and color features from the RGB-D data and fuse them to complete depth map \cite{8869936, 8460184, 10.1007/978-3-030-58601-0_8, Qiu_2019_CVPR, 8917294, Zhang_2018_CVPR, 10.1007/978-3-031-19812-0_13, Lee_2021_CVPR, 2021Decoder}. This approach has shown extraordinary progress compared to the traditional one and is widely applied in recent works. 

However, there are two challenges in the deep learning approach. Firstly, the vanilla convolution operation treats all inputs as valid values \cite{Yu_2019_ICCV}. The raw depth images contain a lot of invalid values, which can defile the latent features extracted by convolution kernels and lead to various visual artifacts during the reconstruction, such as cavities, contradiction, and blurred edges. To ameliorate the limitation, partial convolution has been proposed \cite{Liu_2018_ECCV} to distinguish invalid pixels automatically and calculates the output based only on valid pixels. Moreover, the output pixel will be marked as valid if the receptive field contains at least one valid pixel. This method improves the reliability of the features, but it still has irreconcilable issues. For instance, considering a convolution kernel that covers 3x3 pixels, and no matter how many valid pixels are contained in this region, the kernel outputs are marked equally valid. Nevertheless, the truth is that the confidence of the outputs is totally different in these cases. Going beyond, Gated Convolution (GConv) and De-convolution (De-GConv) \cite{Yu_2019_ICCV} was proposed, which can learn a gating mask via additional convolution kernels to suppress invalid features and strengthen the reliable ones. These operations are workable in extracting features from raw depth images with invalid pixels but fail to handle large missing areas, thus their depth completion results are still not trustworthy. A plausible approach to complete depth images with big holes is taking both color and depth information into account.

Here comes another challenge, using color information for depth completion has both positive and negative effects. Most existing models implement the fusion of color and depth by concatenating latent features directly on the bottleneck of the auto-encoder. However, the involvement of depth-irrelevant color features may mislead the depth prediction results, such as neighboring surfaces in the same color and planes with rich textures. Therefore, a mechanism of screening the interference from the fusion of depth and color features certainly benefits the task of depth completion. Unfortunately, most comparative research has not addressed this problem.

Based on the above observations, we propose a new framework for depth completion based on an UNet-like \cite{10.1007/978-3-319-24574-4_28} architecture, in which the depth and color features are extracted in two parallel encoding branches and then merge into one branch with skip connections in the decoding stage. Specifically, the fusion of the two branches is conducted by stages based on our proposed Attention Guided Gated Convolution (AG-GConv), which learns joint contextual attention from both color and depth values to guide the extraction of depth features. Furthermore, the Attention Guided Skip Connection module is designed to filter out irrelevant color features from the reconstruction of depth. Our main contributions can be summarized as follows:
\begin{itemize}
\item We propose a dual-branch multi-scale encoder-decoder network that combines depth and color features to achieve high-quality completion of the depth image.\vspace{-0.15cm}
\begin{spacing}{0.007}
\end{spacing}
\item An Attention Guided Gated Convolution (AG-GConv) module is proposed to alleviate the adverse impacts of invalid depth values on feature learning.\vspace{-0.15cm}
\begin{spacing}{0.007}
\end{spacing}
\item A new Attention Guided Skip Connection (AG-SC) module is presented to reduce the interference from depth-irrelevant color features to the decoder.\vspace{-0.15cm}
\begin{spacing}{0.007}
\end{spacing}
\item Experimental results indicate that our model outperforms state-of-the-art on three popular benchmarks, including NYU-Depth V2, DIML and SUN RGB-D datasets.\vspace{-0.1cm}
\end{itemize}

\section{Related Works}

\begin{figure*}
  \centering
  \includegraphics[width=\linewidth]{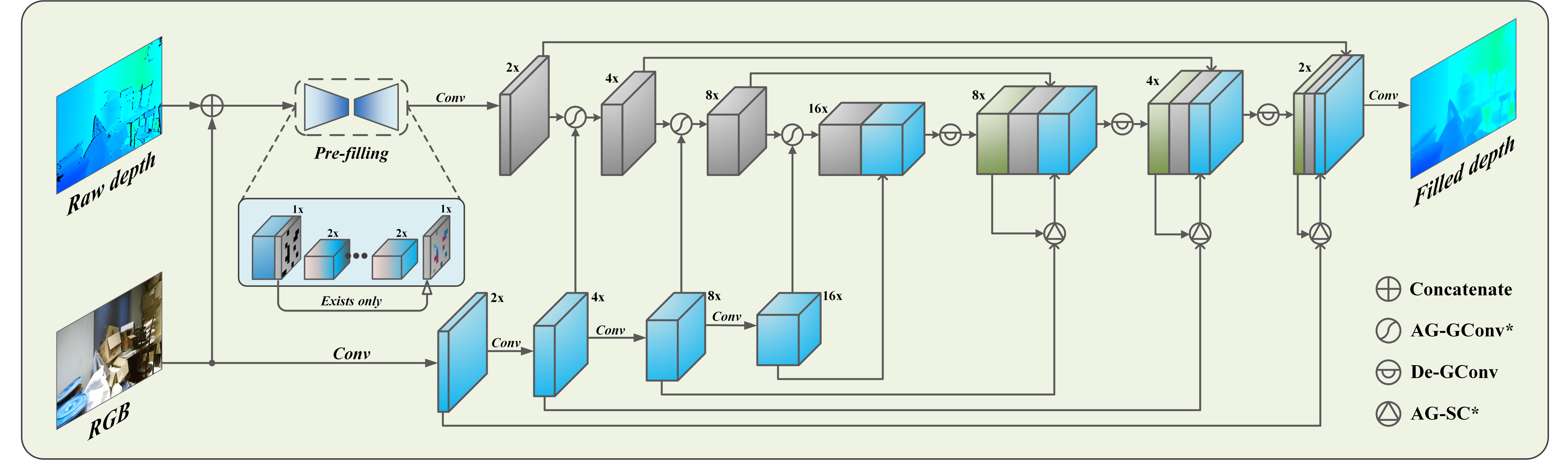}
  \caption{Pipeline of the proposed AGG-Net, where `*' indicates the module we proposed.}
  \label{fig:agg}
\end{figure*}

\textbf{Depth Completion.} In recent years, due to the strong ability of CNN networks on non-linear feature representation, more and more researchers have gradually changed their interests in depth completion from hand-crafted features to deeply learned features. Cheng \etal \cite{8869936} proposed the convolutional spatial propagation network (CSPN), of which the architecture mainly refers to the UNet  \cite{10.1007/978-3-319-24574-4_28} and the ResNet \cite{He_2016_CVPR}. It generates a long-range context through a loop operation, thus reducing the loss of details in the reconstruction. Jaritz \etal \cite{8490955} proposed an auto-encoder framework modified from the famous NASNet \cite{Zoph_2018_CVPR} to obtain a larger receptive field about the input image. Shivakumar \etal \cite{8917294} proposed DFuseNet based on the spatial pyramid pooling module \cite{Zhao_2017_CVPR, Chang_2018_CVPR} to pull contextual cues separately from the intensity image and the depth features, and then fuse them later in the network, which effectively exploits the latent relationship between the two modalities. Huang \etal \cite{2019Indoor} utilizes the self-attention mechanism and boundary consistency schema to improve both the depth boundary and image quality.

However, modification and optimization of the architecture need to be further investigated. Firstly, the UNet and the ResNet adopted by CSPN were initially designed for semantic segmentation and image classification, respectively. Their structures are proved good at catching semantic-aware features while ignoring detailed regression due to the pooling operation, limiting their ability to predict missing depth values accurately. Although the multi-scale structure is adopted on some models like DFuseNet, it is still hard to recover the information lost in the encoding process, as the decoding process entirely depends on the bottleneck layer of the model. Secondly, vanilla convolution cannot handle the invalid area of the depth image, a new fashion of convolutional operation is desirable to facilitate robust and adaptive feature extraction from raw depth images with invalid pixels. Lee \etal \cite{9078070} proposed the CrossGuide network, similar to DFuseNet, and its encoder introduces a sensing module to learn multi-modal features from the RGB-D data. Wang \etal \cite{Wang_2022_CVPR} designed an RGB-D fusion GAN to propagate the features across color and depth features. In this work, we design the AG-GConv modules to capture better depth features in the guidance of color features, especially for handling large missing areas. 
 

\textbf{Contextual Attention.} The attention mechanism has been widely used to highlight the essential part of feature maps in image processing \cite{Hu_2018_CVPR}, \cite{NIPS2015_33ceb07b}. In the task of depth completion, a remarkable work named FuseNet \cite{10.1007/978-3-319-54181-5_14} weights the predicted depth maps with both the global and the local confidence maps based on spatial attention mechanism. A beneficial attempt named DeepLidar conducted by Qiu. \cite{Qiu_2019_CVPR} learns channel attention to guide the combination of color features and the map of surface normal vectors. Senushkin \etal \cite{2021Decoder} proposed a new decoder modulation branch that controls the depth reconstruction via SPADE blocks, which modify the spatial distribution of output signals. We noticed that different feature channels may have different optimal spatial attention maps, which requires a new contextual attention mechanism to arrange different spatial attention patterns for different feature channels. Therefore, we propose the new modules of AG-GConv and AG-SC to modulate the fusion of depth and color features based on joint contextual attention on both the channels and spatial locations.
\section{Method}
The architecture of the AGG-Net proposed for depth image completion is illustrated from overview to details, as shown in Fig. \ref{fig:agg}. The pipeline of the whole model, our proposed Attention-Guided Gated-Convolution (AG-GConv) and Attention-Guided Skip-Connection(AG-SC), and a multi-task loss function are introduced in this section.

\subsection{Architecture}

\textbf{Overview.} The pipeline of our model comprises two successive networks: the pre-filling network and the fine-tuning network. The former takes the raw depth images with the missing area and the corresponding RGB images as inputs and provides a complete depth map by filling all the missing values coarsely through a lightweight autoencoder. The fine-tuning network employs a dual-branch encoder to extract features from both depth and color images. Then it reconstructs the depth images through a multi-scale skip-connected decoder. Furthermore, the proposed AG-GConv and the AG-SC modules are embedded into the encoder and decoder layers, respectively, strengthening the fusion of the two modalities more reasonably and consequently improving the quality of the reconstructed depth images. The whole pipeline will be trained in an end-to-end fashion.

\textbf{Pre-filling.} As plotted in Fig. \ref{fig:agg}, the raw depth image and the RGB image are directly merged into a four-channel multi-modal feature tensor fed into a light auto-encoder with two layers of vanilla convolutions and de-convolutions. The output of the pre-filling network is used to fill the missing area of the raw depth image while keeping the valid depth values unchanged. It is worth noting that the convolutional layers adopt larger-size kernels to ensure that the receptive fields are big enough to cover most invalid areas. Therefore, the pre-filling network can provide a coarsely filled depth image without zero-value pixels.

\textbf{Fine-tuning.} As shown in Fig. \ref{fig:agg}, the fine-tuning network employs a dual-branch UNet-like structure to reconstruct the depth maps through feature encoding and decoding. In each encoding layer, different from vanilla convolution or gated convolution \cite{Yu_2019_ICCV}, the proposed AG-GConv module gates the depth feature tensor with an element-wise mask under the guidance of Contextual Attention (CA) learned based on both depth and color features. In each decoding layer, in addition to the De-GConv module \cite{Yu_2019_ICCV}, the AG-SC module is presented to modulate the skip connection from a color encoding layer to the corresponding depth decoding layer, in which a different attention mechanism is applied to suppress out depth-irrelevant color features.
\subsection{Encoding with AG-GConv}

\begin{figure}[t]
    \centering
  \includegraphics[width=0.95\linewidth]{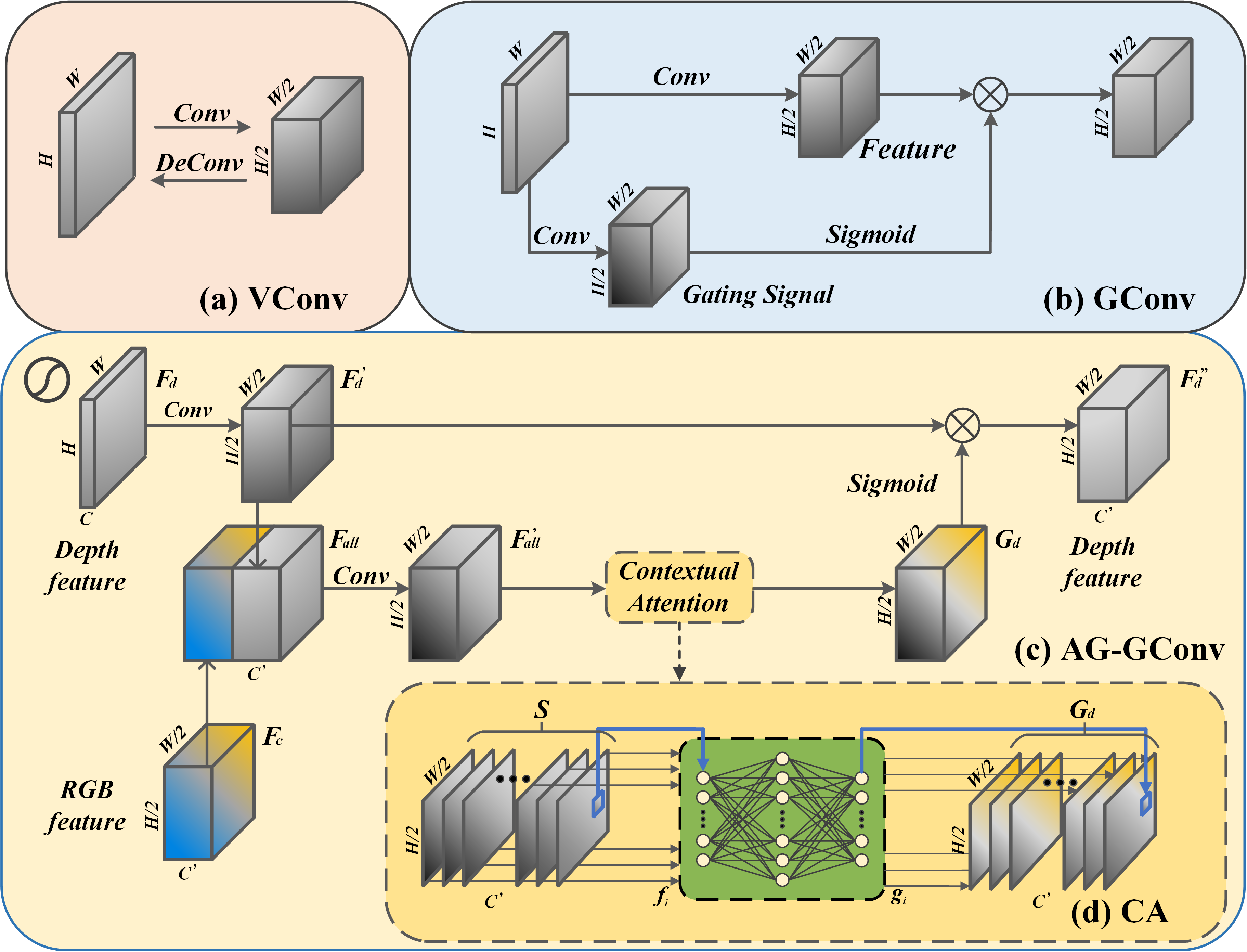}
  \caption{Details of VConv, GConv, AG-GConv and CA.}
  \label{fig:aggc}
\end{figure}

Most traditional encoder networks use multilayer vanilla convolution (VConv) to extract features  \cite{Yu_2019_ICCV}, as shown in Fig. \ref{fig:aggc} (a). Unfortunately, the depth maps provided by the pre-filling network are just filled coarsely. These filled depth values are unreliable and can be transferred or even amplified through the fine-tuning network to pollute the reconstruction results. Gated Convolution \cite{Yu_2019_ICCV} (GConv) provides a better solution to this problem, in which a gating signal is generated to screen out these unreliable features, as shown in Fig. \ref{fig:aggc} (b). However, GConv still has some disadvantages. On the one hand, it only considers the depth features but ignores the valuable information hidden in the color image. On the other hand, it generates the gating signal based on the features of a small receptive field, which weaken its ability to fill large holes.

To overcome the above limitations, we propose a new module named Attention Guided Gated Convolution (AG-GConv), to modulate the depth features under the guidance of contextual attention learned from both depth and color branches. For an AG-GConv module, the input depth and color features are denoted to $F_d$ and $F_c$, respectively. Typically, the size of $F_d$ is $H \times W \times C$, and $F_c$ is in the size of $H/2 \times W/2 \times C'$.

A standard VConv unit contains a 2d-convolutional layer, a batch normalization layer, and a leakyReLU activation layer. At first, as shown in Fig. \ref{fig:aggc} (c), the input depth feature is transformed into $F'_d \in \Re^{H/2 \times W/2 \times C'}$ through two successive VConv units with stride=1 and stride=2 respectively. Then we concatenate $F'_d$ and $F_c$ along the channel axis into a combined feature tensor $F_{all} = [F'_d, F_c]$ whose size is $H/2 \times W/2 \times 2C'$. Next, we deliver $F_{all}$ to a VConv unit with stride=1 to obtain a new feature tensor $F'_{all}$ in the size of $H/2 \times W/2 \times C'$. 

We build the CA module to generate gating signals from $F'_{all}$ by learning the jointly distributed contextual attention between space and channels. At first, it will be split into slices $S=\{ \boldsymbol{s}_i \in \Re^{H/2 \times W/2}| i=1,.., C'\}$ along the channel axis, and then each slice is flattened into a long vector $f_i \in \Re^L$, where $L = H/2 \times W/2$. The proposed CA network consists of two fully-connection layers constructed to learn the global contextual attention, as shown in Fig. \ref{fig:aggc} (d). It is worth noticing that all the slices $S_i$ of $F'_all$ share the same fully-connection layers. Its hidden layer contains $M$ ReLU neurons (typical $M = 4L$), and the output layer contains $L$ Sigmoid neurons. The output vector $\boldsymbol{g}_i$ can be computed as:

\begin{equation}
\setlength{\abovedisplayskip}{2pt}
    \boldsymbol{g}_i = \phi_{ca}(\boldsymbol{f}_i; \boldsymbol{\theta}_{ca})
    \label{eq:gc1}
\setlength{\belowdisplayskip}{2pt}
\end{equation}

\noindent where $\phi_{ca}$ is the mapping function of the CA network, typically ReLU function, with the weight parameters $\boldsymbol{\theta}_{ca}$. The network considers all spatial positions to evaluate the spatial attention for each specific feature slice. Then the output vector 
$\{\boldsymbol{g}_i \in \Re^L| i=1,.., C'\}$ will be reshaped into the size of $H/2 \times W/2$ and be packed into a gating tensor $G_d$ which has the same size as the feature tensor $F'_d$. At last, the output of the AG-GConv module can be obtained by multiplying the depth feature $F'_d$ with the gating tensor $G_{d}$.\vspace{-0.1cm}

\begin{equation}
    {F''_d} = {F'_d} \otimes {G_d}
    \label{eq:gc2}
\end{equation}

\noindent where $\otimes$ denotes the element-wise multiplication between two tensors and the output $F''_d$ of the current AG-GConv module is sent to the following encoding layer as input.


\begin{figure}[h]
~~~~~~~~~~~
\begin{minipage}{0.36\linewidth}
	\vspace{3pt}
	\centerline{\includegraphics[width=\textwidth]{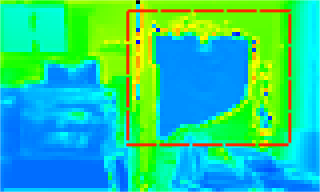}}
	\centerline{(a) \scalebox{0.9}{$F'_d$}}
\end{minipage}
~~~
\begin{minipage}{0.36\linewidth}
	\vspace{3pt}
	\centerline{\includegraphics[width=\textwidth]{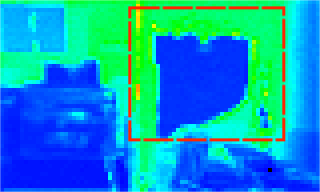}}
	\centerline{(b) \scalebox{0.9}{$F''_d$}}
\end{minipage}
\begin{spacing}{1.2}
\end{spacing}
\caption{Visualization of the depth feature maps (a) $F'_d$ before and (b) $F''_d$ after the modulation of the AG-GConv module.}
\label{fig:fig_explain}
\end{figure}

\textbf{Motivation analysis.} For a raw depth image with large holes, low-level features for pixels inside the hole are unreliable because their neighbors are also invalid. However, vanilla convolutions cannot distinguish these invalid patterns from normal ones as it implements spatial convolution over all the areas in exactly the same fashion. Consequently, these unreliable features will diffuse layer-by-layer and finally degrade the reconstruction results. Although the GConv can restrict the unreliable features with the gating signal, which takes only a small neighborhood into account without considering the large-scale background. That limits the ability to fill large holes in the depth images. Compared to the above two methods, the proposed AG-GConv considers both depth and color features and produces joint contextual attention via the fully connected layers over both space and channels, thus the gating signal is supposed to have more reliability. As shown in Fig. \ref{fig:fig_explain}, most scattered invalid features surrounding the large hole (marked with red boxes) are eliminated when the proposed AG-GConv is applied. It means that the filling process is dominated by reliable features but not invalid ones, which helps to improve the quality and reliability of the reconstructed depth image. 

\subsection{Decoding with AG-SC}

\begin{figure}[t]
  \centering
  \includegraphics[width=\linewidth]{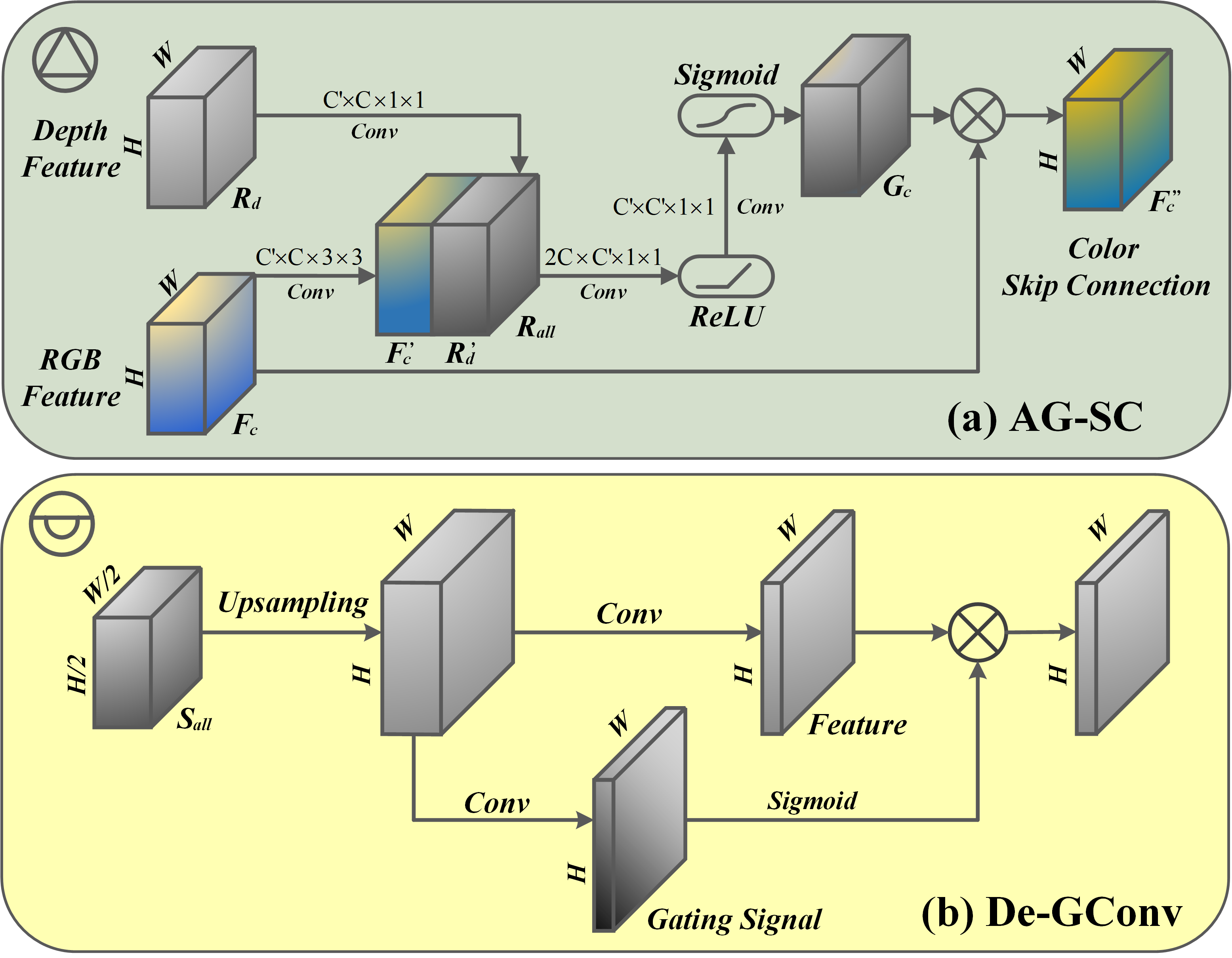}
  \caption{Detailed structure of AG-SC and De-GConv.}
  \label{fig:agsc}
\end{figure}

Most reconstruction networks realize the decoding process with up-sampling and de-convolution. However, the bottleneck between the encoder and the decoder may lead to a severe loss of finer-scale features. Thus, skip connections from the encoder to the decoder are commonly used to make up features in different scales. In addition, color features can be beneficial in predicting depth values, which has been proved by many remarkable works on monocular depth recovery \cite{Zhang_2018_CVPR}. Based on the above analyses, we propose a new decoding scheme as shown in Fig. \ref{fig:agsc}. Each decoding layer collects features from three inputs: the previous layer of the depth branch, skip connections from the depth encoders and the color encoder. Since the depth feature $F_d$ from the stem and the feature from the skip connection have been modulated through De-GConv and AG-GConv modules, the proposed AG-SC module is only used to improve the skip connection from the color branch.

Denoting the features of color skip connection as $F_c$ , and the depth features from the previous layer as $R_d$,  they have the same size of $H\times W\times C$. Through a VConv unit with kernel size $=1\times 1$ and the other one with kernel size $=3\times 3$, $R_d$ and $F_c$ are transformed to $R'_d$ and $F'_c$ , respectively. As shown in Fig. \ref{fig:agsc} (a), the AG-SC module concatenates them into the tensor $R'_{all} = [R'_d; F'_c]$ whose size is $H\times W\times 2C$. Then we use it to learn the gating signal $G_c$ through a VConv unit with an additional ReLU layer and a Sigmoid layer. At last, the output of the AG-SC module can be obtained by implementing the element-wise production according to Eq. \ref{eq:sc1}:

\begin{equation}
    F''_c = F_c \otimes G_c
    \label{eq:sc1}
\end{equation}

We concatenate $R_d$, $F_d$, and $F''_d$ to build a combined feature tensor $S_{all} = [R_d; F_d; F''_c]$, and feed it into the De-GConv module \cite{Yu_2019_ICCV} to produce the output of the current decoder layer, as shown as Fig. \ref{fig:agsc} (b).\vspace{+0.3cm}

\textbf{Motivation analysis.} Considering that there are some potential correlations between the color and the depth images of the same scene, using color features to assist depth prediction is proved to be an effective method for depth completion. However, the correlation between color and depth is complex and uncertain. On the boundaries of an object, color patterns commonly correlate strongly with depth variations. Nevertheless, depth generally keeps constant on a flat surface, while color and texture can change sharply. If the flat surface is a mirror with strong specular reflection or a figured blanket with lower reflectance, a large area of the hole may occur in the corresponding area of the depth image. In these regions, the color information may mislead the depth prediction severely. The AG-SC module is proposed to establish a local attention mechanism by learning the joint distribution of color and depth to suppress the depth-irrelevant color features in the skip connection and reduce their adverse effects on the reconstructed depth images. In the merged tensor $S_{all}$, $R_d$ can provide coarser-scale depth features, $F_d$ from the depth skip connection can supply finer-scale depth features, and the tensor $F''_c$ from the AG-SC can offer filtered color features. It is conceivable that their integration will still introduce some unreliable features. The De-GConv module can partially filter out these destructive features by multiplying the original features with the gating signals, guaranteeing better input for the next decoder layer. In summary, both the proposed AG-SC and the typical De-GConv serve to refine color and depth features for depth completion. An ablative study shows that the integration of the AG-SC and the De-GConv modules does play a positive role in improving the quality of the reconstructed depth images.

\subsection{Loss Functions}

The pipeline is trained in an end-to-end manner under the guidance of the proposed loss function, which consists of two terms, as described in Eq. \ref{eq:loss}:

\begin{equation}
    \mathcal{L}_{\rm{total}} = \lambda_\delta\mathcal{L}_\delta +  \lambda_p\mathcal{L}_p
    \label{eq:loss}
\end{equation}

\noindent where $\mathcal{L}_\delta$ is the Huber loss of the reconstruction error, as defined as Eq. \ref{eq:loss1}:

\begin{equation}
    \mathcal{L}_{\delta} = \sum_{i=1}^M{\sum_{j=1}^N{{\rm huber}\left(\hat{d}_{i, j}, d_{i, j}\right)}}
    \label{eq:loss1}
\end{equation}

\noindent where $\hat{d}_{i, j}$ is the predicted depth value at the location $(i, j)$, and $d_{i, j}$ is the corresponding ground truth, $M$ and $N$ are the height and the width of the reconstructed image, with the value of 5 typically. The Huber loss can provide a robust measure for pixel-wise reconstructed errors, which increases the ability to handle outliers and leads to higher accuracy predictions.

$\mathcal{L}_p$ in Eq. \ref{eq:loss} represents the edge persistence loss, which is defined as:

\begin{equation} 
    \begin{split}
    \mathcal{L}_p = \sum_{i=1}^M {}\sum_{j=1}^N {} \left[ | g_v(\hat{d}_{i, j}) - g_v(d_{i, j}) | \right.\\
    \left. + | g_h(\hat{d}_{i, j}) - g_h(d_{i, j}) |\right]
    \end{split}
    \label{eq:loss2}
\end{equation}

\noindent where the functions $g_v$ and $g_h$ can compute the vertical and the horizontal gradients of depth images, respectively. It is easy to understand that the  $\mathcal{L}_p$ can be used to keep the consistency of boundaries between the reconstructed depth image and the ground truth, which is essential for edge sharping, spatial structure alignment, and the other downstream visual tasks. The combination of the Huber loss $\mathcal{L}_\delta$ and the edge persistence loss $\mathcal{L}_p$ is weighted by $\lambda_\delta$ and $\lambda_p$ jointly, which are designed to balance the emphases on global consistency and local fidelity. 

\section{Experiments}

\subsection{Datasets and Metrics}

We adopted popular datasets, including NYU-Depth V2, DIML, and SUN RGB-D to perform all the experiments. 

\textbf{NYU-Depth V2} \cite{10.1007/978-3-642-33715-4_54} is the most commonly used dataset for depth completion, which contains 1449 sets collected from 464 different indoor scenes. It can be used as a benchmark to evaluate our model and the competing models. We randomly split the dataset into 420 images for training and 1029 for testing. The original image of size $640 \times 480$ is randomly cropped and resized to $324 \times 288$.

\textbf{DIML} \cite{DBLP:journals/corr/abs-1904-10230} is a recently presented dataset consisting of a series of RGB-D frames captured by the Kinect V2 (indoor) or the zed stereo camera (outdoor). Except for typical invalid patterns, it contains many edge shadows and irregular holes, which can be used to evaluate the adaptation ability of the models to various invalid patterns. We only use the indoor part of the dataset, which includes 1609 sets for training and 503 sets for testing. The original images with the size of $512 \times 288$ will be randomly cropped and resized to $320 \times 192$.

\textbf{SUN RGB-D} \cite{Song_2015_CVPR} is a large dataset that contains 10,335 refined RGB-D images captured by four sensors in 19 major scene categories. It serves as the testing for the generalization ability of the models. Following the official scheme, we used 4845 images for training and 4659 for testing. The input images of size $730 \times 530$ were randomly cropped and resized to $384 \times 288$.

\textbf{Metrics.} Three metrics are applied to evaluate the depth completion results: Root Mean Squared Error(RMSE), absolute Relative error(Rel), and $\delta_t$, which is the percentage of predicted depth pixels falling within the thresholds $t=1.10$, $1.25$, $1.25^2$, and $1.25^3$ for finer-grained evaluations.

\subsection{Ablation Studies}
In this section, we report the results of ablation experiments to analyze the effectiveness of our proposed AGG-Net framework. At first, experiments on different pipeline schemes are conducted to validate the contributions of our work and find the best scheme for the proposed model. Then a series of analysis experiments are carried out to optimize some significant hyper-parameters.

\begin{figure*}[h]
    \small
    \centering
	\hspace{10pt}
	\begin{minipage}{0.26\linewidth}
		\vspace{3pt}
		\centerline{\includegraphics[width=\textwidth]{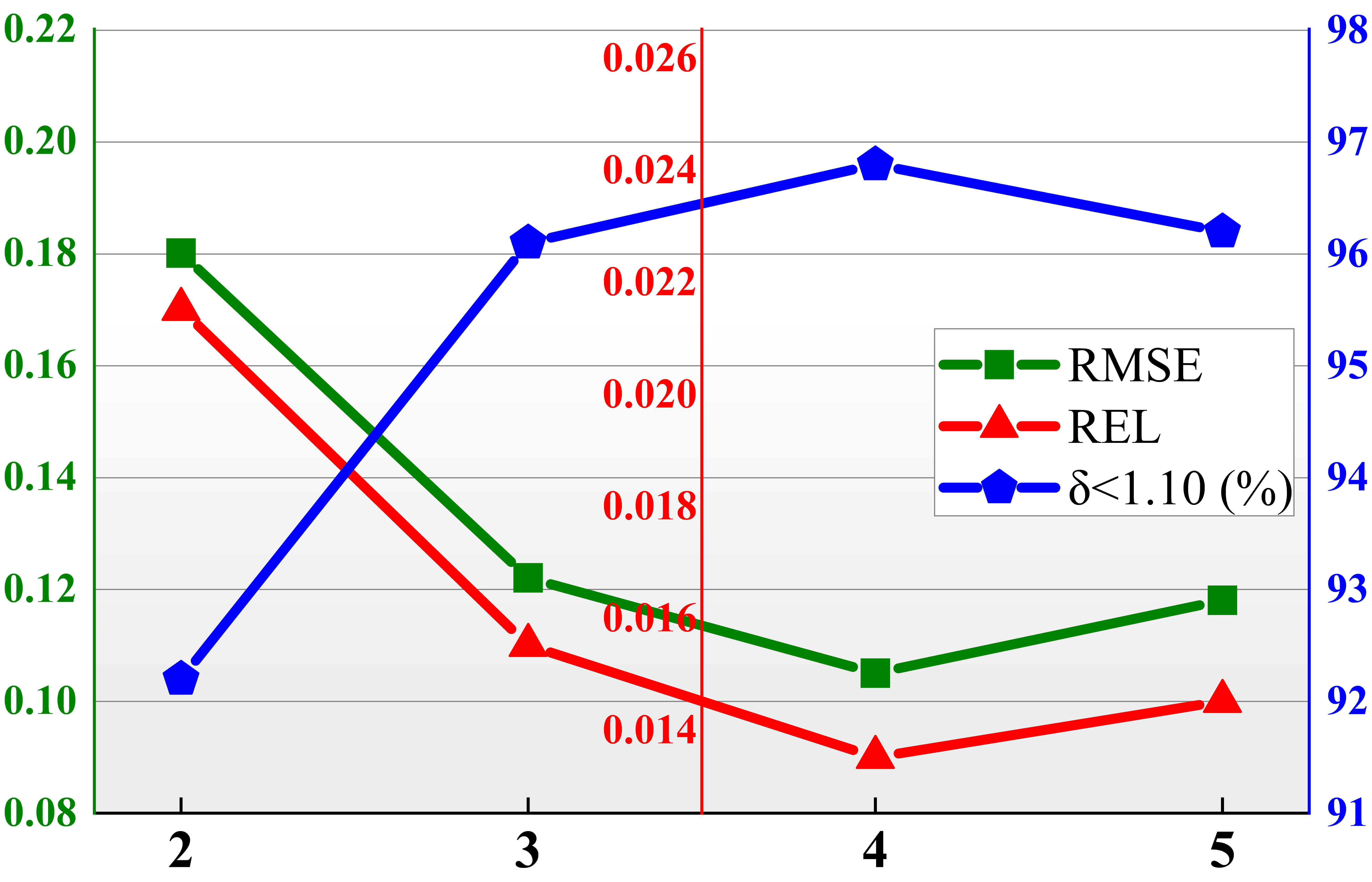}}
		\centerline{(a) Layers' number $m$}
	\end{minipage}
    \centering
    \hspace{10pt}
	\begin{minipage}{0.26\linewidth}
		\vspace{3pt}
		\centerline{\includegraphics[width=\textwidth]{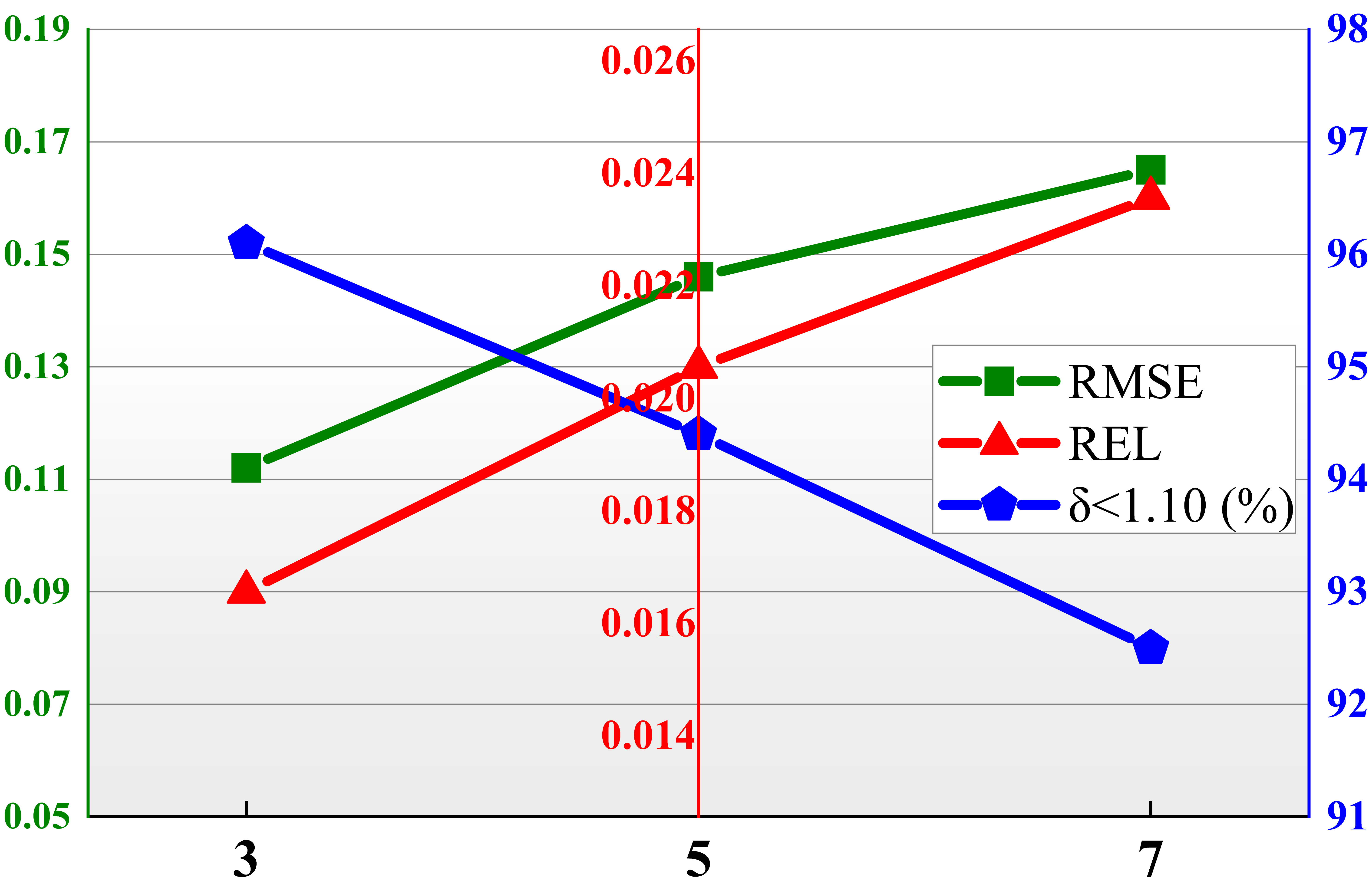}}
		\centerline{(b) Kernel size $k$}
	\end{minipage}
	\centering
	\hspace{10pt}
	\begin{minipage}{0.26\linewidth}
		\vspace{3pt}
		\centerline{\includegraphics[width=\textwidth]{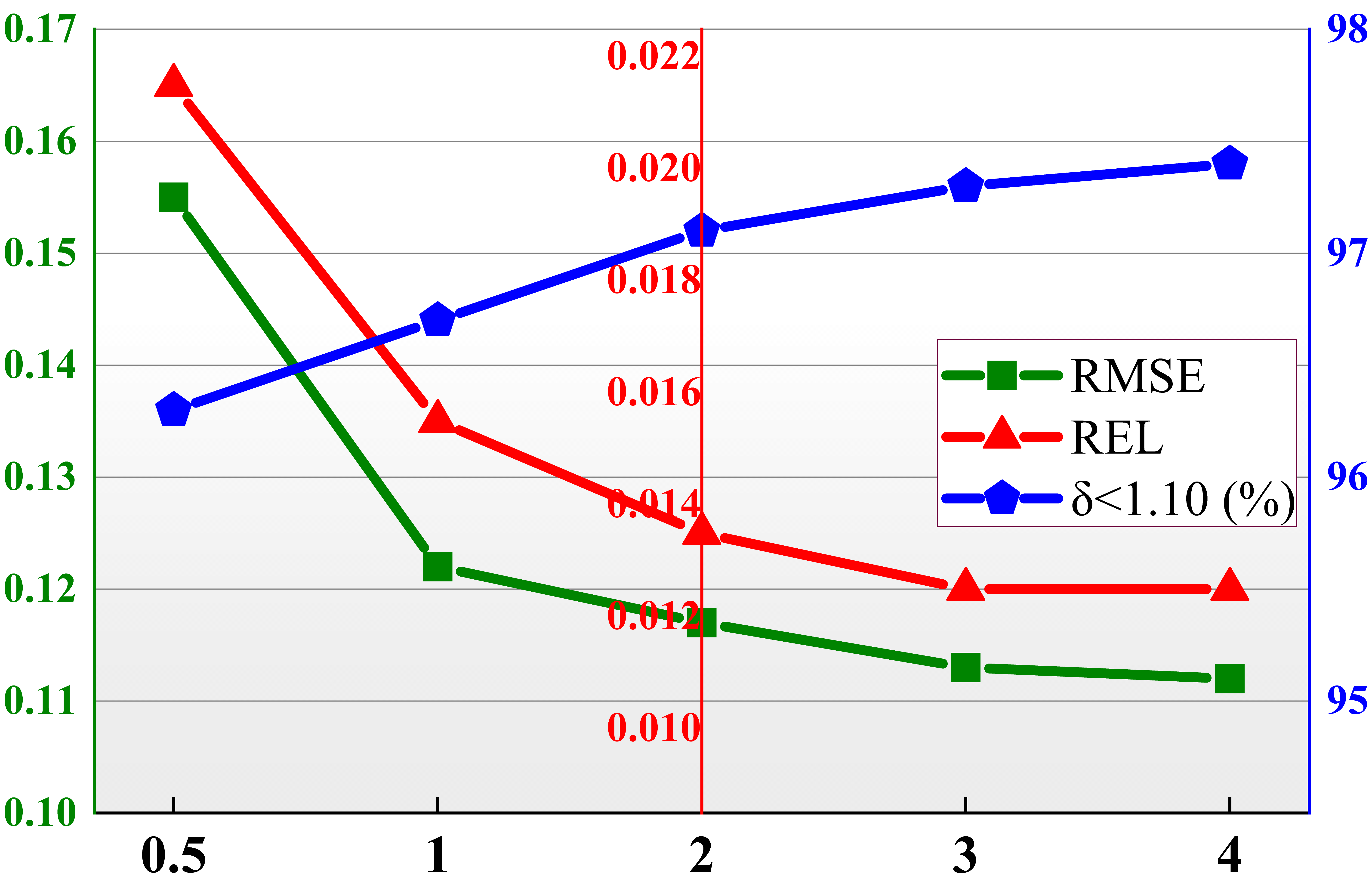}}
		\centerline{(c) Ratio $r$ of CA}
	\end{minipage}
    \begin{spacing}{1.2}
    \end{spacing}
	\caption{Ablation study of parameters. RSME (left axis, lower the better), REL (middle axis, lower the better) and $\delta < 1.10$ (right axis, higher the better) w.r.t. (a) Layers' number $m$, (b) Kernel size $k$, and (c) Ratio $r$ for CA.}
	\label{fig_ablations}
\end{figure*}

\textbf{Settings.} Our framework is implemented in Pytorch and trained using the SDG optimizer. We use batch sizes of 8 for both training and testing experiments. The initial learning rate is $\eta=10^{-2}$, and it falls to 30\% on a plateau until the minimum value of $\eta=10^{-4}$. The momentum term is 0.95, and the weight decay term is $10^{-4}$. The model's setting in the parameters ablation study is $m = 3$, $k = 3$, $r = 1$, whereas the best setting in other experiments is $m = 4$, $k = 3$, $r = 4$. The weights of the loss in Eq. \ref{eq:loss} are empirically set as $\lambda_\delta = 0.7$, $\lambda_p=0.3$. All the corresponding models are trained on the NYU-Depth v2 for 120 epochs, and the three metrics RMSD, Rel, and $\delta_t$ are reported for evaluation. 

\begin{table}[h]
  \centering
  \setlength{\tabcolsep}{2pt}
  \scalebox{0.8}{
  \begin{tabular}{c|ccccc|ccc}
    \hline
    Scheme & Fusion & Pre. & GC. & AG-GC. & AG-SC & RMSE$\downarrow$ &
    Rel$\downarrow$ & $\delta_{1.10}\uparrow$ \\
    \hline\hline
    A & None& $\checkmark$ &&&& 0.136 & 0.026 & 96.6\\
    B & Concat. & $\checkmark$ &&&& 0.122 & 0.024 & 97.0 \\
    \hline
    C & Concat. & $\checkmark$ & $\checkmark$ &&& 0.115 & 0.018 & 97.5\\
    D & Guided & $\checkmark$ &&$\checkmark$&& 0.105 & 0.016 & 97.6\\
    \hline
    E & Concat. & $\checkmark$ & $\checkmark$ &&$\checkmark$& 0.103 & 0.016 & 97.8\\
    F & Guided &&& $\checkmark$ & $\checkmark$ & 0.105 & 0.016 & 97.4\\
    \textbf{G} & \textbf{Guided} & \textbf{$\checkmark$} && \textbf{$\checkmark$} & \textbf{$\checkmark$} & \textbf{0.092} & \textbf{0.014} & \textbf{98.3}\\
    \hline
  \end{tabular}
  }
  \vspace{+0.3cm}
  \caption{Ablation study results for different schemes of the pipeline. `Fusion' represents the mode of the fusion between depth and color.}
  \label{tab:ablation}
\end{table}

\textbf{On pipeline.} The pipeline of the baseline framework is the UNet-like architecture, containing only the branch of depth with vanilla convolution and skip connections (Scheme A in Tab. \ref{tab:ablation}). Then a group of ablation experiments is implemented on different schemes of the baseline model and the proposed modules. In the scheme B, the fusion of depth and color is carried out by concatenating their feature tensors directly at each encoder layer. The results show that the addition of color features does benefit a lot to depth completion. The scheme C adopts the GConv and De-GConv modules instead of vanilla convolution and de-convolution, which performs better because the gating signals can help filter out unreliable features. In the scheme D, we replace the GConv with our proposed AG-GConv. All the three metrics are improved significantly because of the introduction of contextual attention. In the scheme E, we replace the color skip connection with our proposed AG-SC module while keeping the GConv and De-GConv modules. The improved performance proves the contribution of the proposed AG-SC. In the scheme F, we remove the Pre-filling module to investigate its effectiveness. At last, we integrate the Pre-filling, AG-GConv and the AG-SC together in the scheme G, and all the three measures reach their top scores, which proves that all the three modules help the model enhance the performance. 

All the above results demonstrate that both the proposed AG-GConv and the AG-SC contribute to the depth completion task. The advantages of the AG-GConv over the GConv originate from the fact that the former can learn contextual attention for gating signals based on both depth and color features, while the latter uses only the depth features with local attention. The improvements caused by the proposed AG-SC module indicate that purifying color features with local attention is helpful to the reconstruction of depth images. Moreover, their combination can further improve the final performance of our model. The accuracy of the optimal scheme increases significantly compared with the baseline, and the RMSE and Rel values reduce by around 32.4\% and 46.2\%, respectively. According to the results, the optimal scheme of our model is the dual-branch UNet-like architecture embedded with the AG-GConv and the AG-SC modules, as shown in Fig. \ref{fig:agg}.

\textbf{On layers.} The encoder-decoder architecture can realize multi-scale feature extraction for image reconstruction, of which the scale distribution heavily depends on the layers' number $m$ of the encoder and the decoder. The curves of RMSE, REL, and $\delta_{1.10}$ are plotted against different layers' numbers $m=2,3,4$ and $5$, as shown in Fig. \ref{fig_ablations} (a). It can be seen that all three measures reach saturation at $m=4$, which is consequently adopted as the default setting of the layers' number.

\textbf{On kernel size.} The size of convolutional kernels decides the receptive fields of neurons and gating signals, and it influences the final performance dramatically. As shown in Fig. \ref{fig_ablations} (b), all the three indexes get their tops at $k=3$. It is because those larger receptive fields are more likely to involve irrelevant depth pixels, especially at the boundaries. Therefore, we set the kernel size $k=3$ for most convolutional layers.

\textbf{On hidden layer of CA.} The number of neurons in the Contextual Attention module's hidden layer significantly influences the capacity of the proposed AG-GConv module. The ratio $r$ between the input neurons and the hidden-layer neurons can be regarded as an essential parameter to optimize the final performance of the model. As shown in Fig. \ref{fig_ablations} (c), where the curves of the three measures are plotted vs. $r=2, 3, 4, 5$, the performance is saturated when the ratio increases to $r=4$. That is why we build the CA module with $4\times H\times W$ hidden-layer neurons.

\begin{table}[h]
  \centering
  \setlength{\tabcolsep}{2pt}
  \scalebox{0.8}{
  \begin{tabular}{c|ccc}
    \hline
    Loss & RMSE$\downarrow$ & Rel$\downarrow$ & $\delta_{1.10}\uparrow$\\
    \hline\hline
    $\mathcal{L}_{mse}$ & 0.104 & 0.016 & 97.9\\
    $\mathcal{L}_\delta$ & 0.100 & 0.015 & 98.0\\
    \textbf{$\mathcal{L}_\delta+\mathcal{L}_p$} & \textbf{0.092} & \textbf{0.014} & \textbf{98.3}\\
    \hline
  \end{tabular}
  }
  \vspace{+0.3cm}
  \caption{Ablation study results on different loss functions. }
  \label{tab:ablation_loss}
\end{table}

\textbf{On loss function.} Unlike the traditional reconstruction loss based on the Mean Square Error, we introduce the Huber loss to accommodate the outlier pixels in the reconstructed depth images and present the edge persistence loss to emphasize the boundaries of different surfaces. As shown in Tab. \ref{tab:ablation_loss}, the Huber loss obtains slight improvements on all three measures compared with the MSE loss, while adding the edge persistence loss dramatically improves the performance.

\begin{figure*}[t]
\centering
\includegraphics[width=0.13\linewidth]{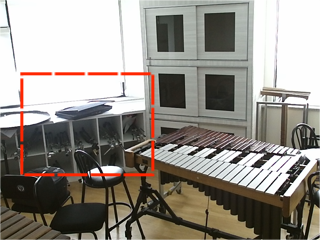}\vspace{1pt}
\includegraphics[width=0.13\linewidth]{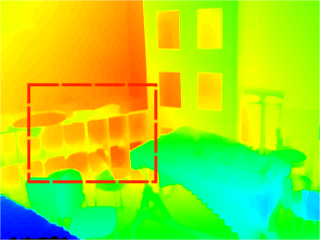}\vspace{1pt}
\includegraphics[width=0.13\linewidth]{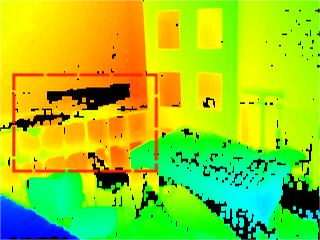}\vspace{1pt}
\includegraphics[width=0.13\linewidth]{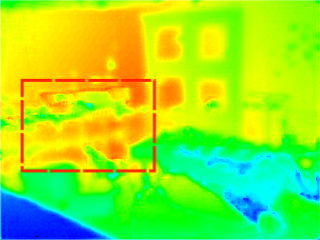}\vspace{1pt}
\includegraphics[width=0.13\linewidth]{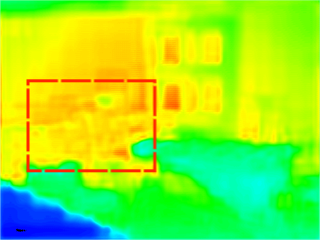}\vspace{1pt}
\includegraphics[width=0.13\linewidth]{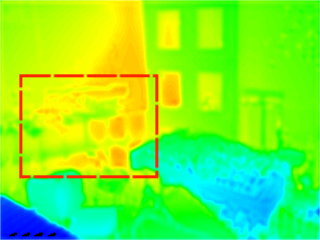}\vspace{1pt}
\includegraphics[width=0.13\linewidth]{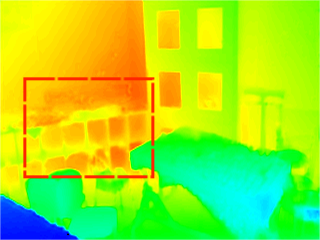}\vspace{1pt}
\begin{spacing}{0.5}
\end{spacing}
\includegraphics[width=0.13\linewidth]{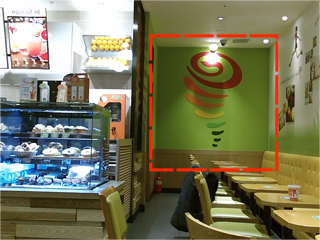}\vspace{1pt}
\includegraphics[width=0.13\linewidth]{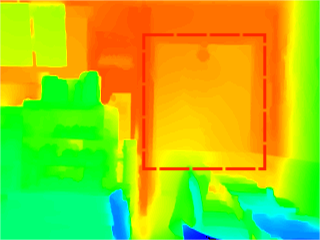}\vspace{1pt}
\includegraphics[width=0.13\linewidth]{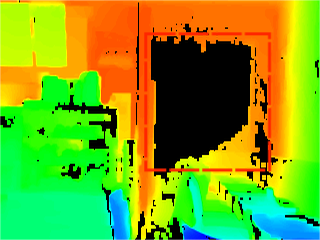}\vspace{1pt}
\includegraphics[width=0.13\linewidth]{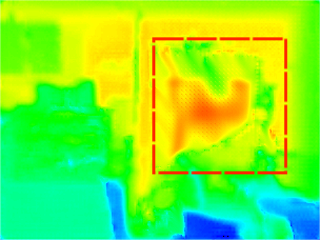}\vspace{1pt}
\includegraphics[width=0.13\linewidth]{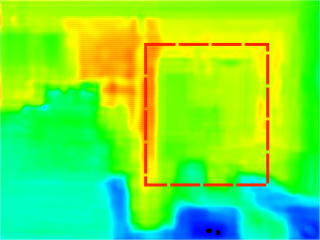}\vspace{1pt}
\includegraphics[width=0.13\linewidth]{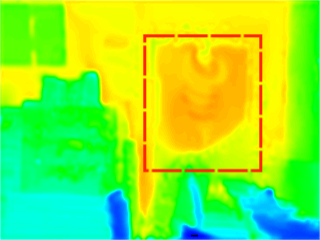}\vspace{1pt}
\includegraphics[width=0.13\linewidth]{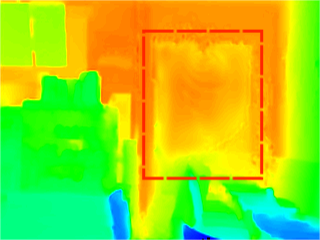}\vspace{1pt}
\begin{spacing}{0.5}
\end{spacing}
\includegraphics[width=0.13\linewidth]{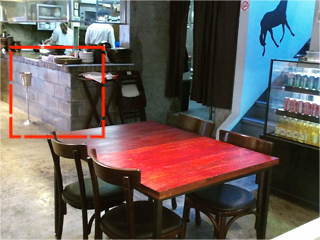}\vspace{1pt}
\includegraphics[width=0.13\linewidth]{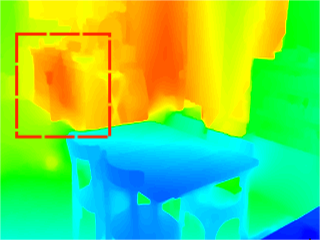}\vspace{1pt}
\includegraphics[width=0.13\linewidth]{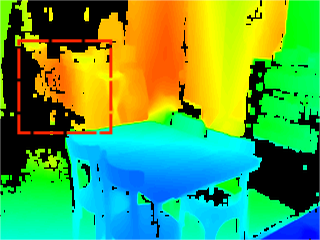}\vspace{1pt}
\includegraphics[width=0.13\linewidth]{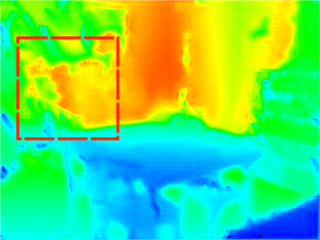}\vspace{1pt}
\includegraphics[width=0.13\linewidth]{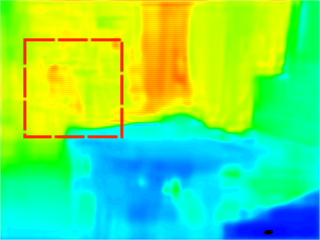}\vspace{1pt}
\includegraphics[width=0.13\linewidth]{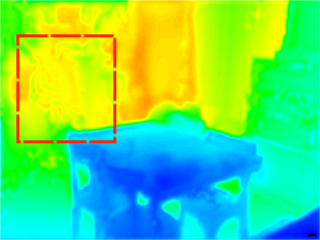}\vspace{1pt}
\includegraphics[width=0.13\linewidth]{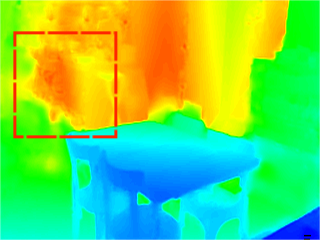}\vspace{1pt}
\begin{spacing}{0.5}
\end{spacing}
\begin{minipage}{0.13\linewidth}
\centerline{\includegraphics[width=\textwidth]{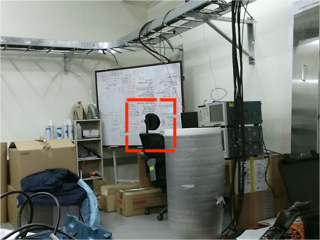}}
\centerline{RGB}\vspace{1pt}
\end{minipage}
\begin{minipage}{0.13\linewidth}
\centerline{\includegraphics[width=\textwidth]{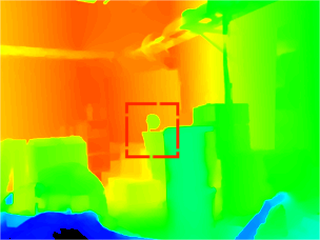}}
\centerline{GT}\vspace{1pt}
\end{minipage}
\begin{minipage}{0.13\linewidth}
\centerline{\includegraphics[width=\textwidth]{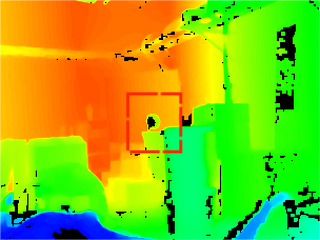}}
\centerline{Raw}\vspace{1pt}
\end{minipage}
\begin{minipage}{0.13\linewidth}
\centerline{\includegraphics[width=\textwidth]{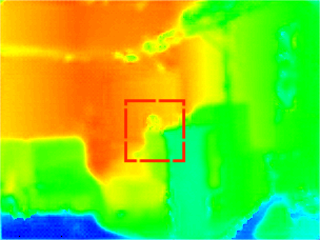}}
\centerline{CSPN*}\vspace{1pt}
\end{minipage}
\begin{minipage}{0.13\linewidth}
\centerline{\includegraphics[width=\textwidth]{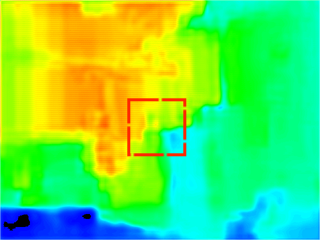}}
\centerline{DFuseNet*}\vspace{1pt}
\end{minipage}
\begin{minipage}{0.13\linewidth}
\centerline{\includegraphics[width=\textwidth]{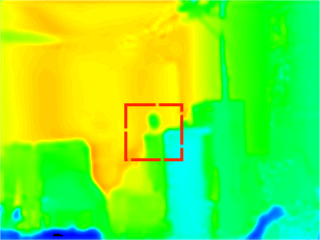}}
\centerline{DM-LRN}\vspace{1pt}
\end{minipage}
\begin{minipage}{0.13\linewidth}
\centerline{\includegraphics[width=\textwidth]{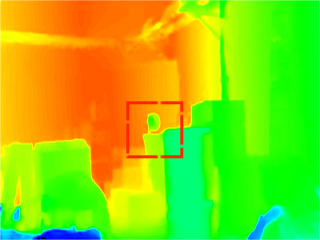}}
\centerline{Ours}\vspace{1pt}
\end{minipage}
\begin{spacing}{1.6}
\end{spacing}
\caption{Depth completion comparison results with different methods on DIML.}
\label{fig:comparison}
\end{figure*}

\subsection{Comparison to State-of-the-art}

To validate the performance of our proposed AGG-Net, we compare our method against the classic bilateral filtering method \cite{1467423}, and various latest SOTA work \cite{8460184, 8869936, 8917294, 10.1007/978-3-030-58601-0_8, Wang_2022_CVPR, Qiu_2019_CVPR} based on deep learning. Our model is implemented by the optimal scheme and parameters based on the ablation study results. The performance measurements of the competing models are collected from their original publications or obtained by applying them to the above three datasets according to their default settings. All the methods are trained based on the raw depth map and tested in exactly the same protocol. The results are shown in Tab. \ref{tab:comparison}.

\begin{table}[h]
  \centering
  \setlength{\tabcolsep}{2pt}
  \scalebox{0.8}{
  \begin{tabular}{c|c|c|c|ccc}
    \hline
    Benchmark & Method & RMSE$\downarrow$ & Rel$\downarrow$ & $\delta_{1.25}\uparrow$ & $\delta_{1.25^2}\uparrow$ & $\delta_{1.25^3}\uparrow$ \\
    \hline\hline
    \multirow{7}{*}{(a) NYU.} & Bilateral \cite{1467423} & 0.532 & 0.132 & 85.1 & 93.5 & 95.9 \\
    ~ & Sparse2Dense* \cite{8460184} & 0.230 & 0.054 & 94.5 & 97.3 & 98.9\\
    ~ & CSPN* \cite{8869936} & 0.173 & 0.020 & 96.3 & 98.6 & 99.5\\
    ~ & DfuseNet* \cite{8917294} & 0.156 & 0.016 & 98.8 & 99.7 & 99.9\\
    ~ & DM-LRN \cite{2021Decoder} & 0.205 & 0.014 & 98.8 & 99.6 & 99.9\\
    ~ & RDF-GAN \cite{Wang_2022_CVPR} & 0.139 & \textbf{0.013} & 98.7 & 99.6 & 99.9\\
    ~ & \textbf{Ours} & \textbf{0.092} & 0.014 & \textbf{99.4} & \textbf{99.9} & \textbf{100.0}\\
    \hline
    \multirow{5}{*}{(b) DIML} & Bilateral \cite{1467423} & 0.636 & 0.189 & 83.0 & 88.8 & 92.4 \\
    ~ & CSPN* \cite{8869936} & 0.162 & 0.033 & 96.1 & 98.7 & 99.6\\
    ~ & DfuseNet* \cite{8917294} & 0.143 & 0.023 & 98.4 & 99.4 & 99.9\\
    ~ & DM-LRN \cite{2021Decoder} & 0.149 & 0.015 & 99.0 & 99.6 & 99.9\\
    ~ & \textbf{Ours} & \textbf{0.078} & \textbf{0.011} & \textbf{99.6} & \textbf{99.9} & \textbf{100.0}\\
    \hline
    \multirow{6}{*}{(c) SUN.} & Sparse2Dense* \cite{8460184} & 0.329 & 0.074 & 93.9 & 97.0 & 98.1 \\
    ~ & CSPN* \cite{8869936} & 0.295 & 0.137 & 95.6 & 97.5 & 98.4\\
    ~ & DeepLidar \cite{Qiu_2019_CVPR} & 0.279 & 0.061 & 96.9 & 98.0 & 98.4\\
    ~ & DM-LRN \cite{2021Decoder} & 0.267 & 0.063 & 97.6 & 98.2 & 98.7\\
    ~ & RDF-GAN \cite{Wang_2022_CVPR} & 0.255 & 0.059 & 96.9 & 98.4 & 99.0\\
    ~ & \textbf{Ours} & \textbf{0.152} & \textbf{0.038} & \textbf{98.5} & \textbf{99.0} & \textbf{99.4}\\
    \hline
  \end{tabular}
  }
  \vspace{+0.3cm}
  \caption{Quantitative comparison results with other methods on (a) NYU-Depth V2, (b) DIML, and (c) SUN RGB-D. `*' indicates that the method is originally designed for sparse depth completion.}
  \label{tab:comparison}
\end{table}

\textbf{On NYU-Depth V2.} On the most popular benchmark NYU-Depth V2, the deep learning models \cite{8460184, 8869936, 8917294, 2021Decoder, Wang_2022_CVPR} are obviously superior to the traditional Bilateral method \cite{1467423} because deep learning is more potent than traditional image processing techniques in dealing with depth completion. Among all the learning-based methods, our model scores the best and wins the second by a large margin of 33.8\% (0.092 vs. 0.139) in RMSE. Our model takes the top scores of $\delta_t$, and reaches 100\% at the threshold $t=1.25^3$. For the index of Rel, our model achieves almost the top score with just $0.001$ point behind. Considering that the NYU-Depth V2 dataset is the most widely used benchmark in this field, the corresponding results give confident evidence that our model outperforms most existing SOTA works in the average performance. The overall advantage of our proposed AGG-Net derives from the proposed AG-GConv and AG-SC modules.

\textbf{On DIML.} As a new dataset, the DIML is characterized by some novel patterns of invalid areas, including spotted mass, edge shadows, and large irregular holes. Our model is compared with the Bilateral and the other three SOTA methods \cite{8869936, 8917294, 2021Decoder} on the DIML dataset using the same testing protocol. The results show that our model further improves the Rel score while maintaining or expanding the advantages on RMSE and $\delta_t$. We also display several typical images in the DIML datasets and the corresponding completion results in Fig. \ref{fig:comparison}. It can be clearly seen that, especially marked with the red box, our results have more vivid details, sharper edges, and higher consistency to the GT compared to the competing models, even for those very challenging cases with large irregular holes and dense speckles. These results demonstrate that our model has a more robust adaptation to various patterns of depth missing, which is believed to come with the better learning ability of the proposed AG-GConv and AG-SC modules.

\textbf{On SUN RGB-D.} The experiments on the large-scale dataset SUN RGB-D are conducted to evaluate the generalization ability of our model. As shown in Tab. \ref{tab:comparison}, our model vastly outperforms the other SOTA methods in all the metrics. Especially compared to the latest SOTA method RDF-GAN which ranks the second, our model improves the RMSE score by 50\% (0.128 vs. 0.256) and the Rel score by around 40\% (0.035 vs. 0.059), respectively. It is considered that the multi-scale architecture, the fusion of depth and color, the global contextual attention in the AG-GConv, and the local attention of the AG-SC module jointly enhance the generalization ability of our model on a wide variety of scenes.

All the above experimental results prove that our method provides a new strong baseline for depth completion. The proposed architecture, AG-GConv and AG-SC modules significantly contribute to the promotion of the AGG-Net model, and make it outperform most existing depth completion methods on the popular benchmarks and metrics.

\section{Conclusion}
Our proposed Attention-Guided Gated-convolutional network (AGG-Net) provides a more robust baseline for depth completion tasks. It accommodates feature extraction and depth reconstruction within a dual-branch UNet-like architecture embedded with the proposed AG-GConv and AG-SC modules. The proposed AG-GConv can modulate the fusion of depth and color features by learning global contextual attention. In addition, the proposed AG-SC contributes to depth reconstruction by highlighting important color features while suppressing depth-irrelevant ones. The experimental results demonstrate that our proposed AGG-Net outperforms the state-of-the-art methods on the popular benchmarks NYU-Depth V2, DIML, and SUN RGB-D.

\section{Acknowledgments}
This research was supported by the Fundamental Research Funds for the Central Universities (N2104027); Innovation Fund of Chinese Universities Industry-University-Research (2020HYA06003); National Natural Science Foundation of China (62202087, U22A2063); National Key Research and Development Project of China (No.2022YFF0902401); Guangdong Basic and Applied Basic Research Foundation (2021B1515120064, 2021A1515110761); Major Program of National Natural Science Foundation of China (71790614); the 111 Project (B16009).

{\small
\bibliographystyle{ieee_fullname}
\bibliography{egbib}

\begin{thebibliography}{10}\itemsep=-1pt

\bibitem{378185}
B. Bascle and R. Deriche.
\newblock Stereo matching, reconstruction and refinement of 3d curves using
  deformable contours.
\newblock In {\em 1993 (4th) International Conference on Computer Vision},
  pages 421--430, 1993.

\bibitem{1467423}
A. Buades, B. Coll, and J.-M. Morel.
\newblock A non-local algorithm for image denoising.
\newblock In {\em IEEE Computer Society Conference on Computer Vision and
  Pattern Recognition (CVPR)}, pages 60--65 vol. 2, 2005.

\bibitem{Chang_2018_CVPR}
Jia-Ren Chang and Yong-Sheng Chen.
\newblock Pyramid stereo matching network.
\newblock In {\em Proceedings of the IEEE Conference on Computer Vision and
  Pattern Recognition (CVPR)}, 2018.

\bibitem{Chen_2015_ICCV}
Chenyi Chen, Ari Seff, Alain Kornhauser, and Jianxiong Xiao.
\newblock Deepdriving: Learning affordance for direct perception in autonomous
  driving.
\newblock In {\em Proceedings of the IEEE International Conference on Computer
  Vision (ICCV)}, 2015.

\bibitem{8869936}
Xinjing Cheng, Peng Wang, and Ruigang Yang.
\newblock Learning depth with convolutional spatial propagation network.
\newblock {\em IEEE Transactions on Pattern Analysis and Machine Intelligence},
  pages 2361--2379, 2020.

\bibitem{DBLP:journals/corr/abs-1904-10230}
Jaehoon Cho, Dongbo Min, Youngjung Kim, and Kwanghoon Sohn.
\newblock A large {RGB-D} dataset for semi-supervised monocular depth
  estimation.
\newblock {\em CoRR}, 2019.

\bibitem{Ferstl_2013_ICCV}
David Ferstl, Christian Reinbacher, Rene Ranftl, Matthias Ruether, and Horst
  Bischof.
\newblock Image guided depth upsampling using anisotropic total generalized
  variation.
\newblock In {\em Proceedings of the IEEE International Conference on Computer
  Vision (ICCV)}, 2013.

\bibitem{10.1007/978-3-319-54181-5_14}
Caner Hazirbas, Lingni Ma, Csaba Domokos, and Daniel Cremers.
\newblock Fusenet: Incorporating depth into semantic segmentation via
  fusion-based cnn architecture.
\newblock In {\em Computer Vision -- ACCV 2016}, pages 213--228, 2017.

\bibitem{He_2016_CVPR}
Kaiming He, Xiangyu Zhang, Shaoqing Ren, and Jian Sun.
\newblock Deep residual learning for image recognition.
\newblock In {\em Proceedings of the IEEE Conference on Computer Vision and
  Pattern Recognition (CVPR)}, 2016.

\bibitem{Hu_2018_CVPR}
Jie Hu, Li Shen, and Gang Sun.
\newblock Squeeze-and-excitation networks.
\newblock In {\em Proceedings of the IEEE Conference on Computer Vision and
  Pattern Recognition (CVPR)}, 2018.

\bibitem{2019Indoor}
Yu-Kai Huang, Tsung-Han Wu, Yueh-Cheng Liu, and Winston~H. Hsu.
\newblock Indoor depth completion with boundary consistency and self-attention.
\newblock In {\em 2019 IEEE/CVF International Conference on Computer Vision
  Workshop (ICCVW)}, pages 1070--1078, 2019.

\bibitem{NIPS2015_33ceb07b}
Max Jaderberg, Karen Simonyan, Andrew Zisserman, and koray kavukcuoglu.
\newblock Spatial transformer networks.
\newblock In {\em Advances in Neural Information Processing Systems}, 2015.

\bibitem{8490955}
Maximilian Jaritz, Raoul~De Charette, Emilie Wirbel, Xavier Perrotton, and
  Fawzi Nashashibi.
\newblock Sparse and dense data with cnns: Depth completion and semantic
  segmentation.
\newblock In {\em 2018 International Conference on 3D Vision (3DV)}, pages
  52--60, 2018.

\bibitem{Lee_2021_CVPR}
Byeong-Uk Lee, Kyunghyun Lee, and In~So Kweon.
\newblock Depth completion using plane-residual representation.
\newblock In {\em Proceedings of the IEEE/CVF Conference on Computer Vision and
  Pattern Recognition (CVPR)}, pages 13916--13925, 2021.

\bibitem{9078070}
Sihaeng Lee, Janghyeon Lee, Doyeon Kim, and Junmo Kim.
\newblock Deep architecture with cross guidance between single image and sparse
  lidar data for depth completion.
\newblock {\em IEEE Access}, pages 79801--79810, 2020.

\bibitem{10.1007/BFb0028368}
S.~Z. Li.
\newblock Markov random field models in computer vision.
\newblock In {\em Proceedings of the European Conference on Computer Vision
  (ECCV)}, pages 361--370, 1994.

\bibitem{Liu_2018_ECCV}
Guilin Liu, Fitsum~A. Reda, Kevin~J. Shih, Ting-Chun Wang, Andrew Tao, and
  Bryan Catanzaro.
\newblock Image inpainting for irregular holes using partial convolutions.
\newblock In {\em Proceedings of the European Conference on Computer Vision
  (ECCV)}, 2018.

\bibitem{8460184}
Fangchang Ma and Sertac Karaman.
\newblock Sparse-to-dense: Depth prediction from sparse depth samples and a
  single image.
\newblock In {\em 2018 IEEE International Conference on Robotics and Automation
  (ICRA)}, pages 4796--4803, 2018.

\bibitem{10.1007/978-3-030-58601-0_8}
Jinsun Park, Kyungdon Joo, Zhe Hu, Chi-Kuei Liu, and In So~Kweon.
\newblock Non-local spatial propagation network for depth completion.
\newblock In {\em Proceedings of the European Conference on Computer Vision
  (ECCV)}, pages 120--136, 2020.

\bibitem{Qiu_2019_CVPR}
Jiaxiong Qiu, Zhaopeng Cui, Yinda Zhang, Xingdi Zhang, Shuaicheng Liu, Bing
  Zeng, and Marc Pollefeys.
\newblock Deeplidar: Deep surface normal guided depth prediction for outdoor
  scene from sparse lidar data and single color image.
\newblock In {\em Proceedings of the IEEE/CVF Conference on Computer Vision and
  Pattern Recognition (CVPR)}, 2019.

\bibitem{10.1007/978-3-319-24574-4_28}
Olaf Ronneberger, Philipp Fischer, and Thomas Brox.
\newblock U-net: Convolutional networks for biomedical image segmentation.
\newblock In {\em Medical Image Computing and Computer-Assisted Intervention --
  MICCAI 2015}, pages 234--241, 2015.

\bibitem{1307213}
K. Sabe, M. Fukuchi, J.-S. Gutmann, T. Ohashi, K. Kawamoto, and T. Yoshigahara.
\newblock Obstacle avoidance and path planning for humanoid robots using stereo
  vision.
\newblock In {\em IEEE International Conference on Robotics and Automation,
  2004. Proceedings. ICRA '04. 2004}, pages 592--597 Vol.1, 2004.

\bibitem{2021Decoder}
D. Senushkin, I. Belikov, and A. Konushin.
\newblock Decoder modulation for indoor depth completion.
\newblock pages 2181--2188, 2021.

\bibitem{doi:10.1137/S0036144598347059}
J.~A. Sethian.
\newblock Fast marching methods.
\newblock {\em SIAM Review}, pages 199--235, 1999.

\bibitem{8917294}
Shreyas~S. Shivakumar, Ty Nguyen, Ian~D. Miller, Steven~W. Chen, Vijay Kumar,
  and Camillo~J. Taylor.
\newblock Dfusenet: Deep fusion of rgb and sparse depth information for image
  guided dense depth completion.
\newblock In {\em 2019 IEEE Intelligent Transportation Systems Conference
  (ITSC)}, pages 13--20, 2019.

\bibitem{10.1007/978-3-642-33715-4_54}
Nathan Silberman, Derek Hoiem, Pushmeet Kohli, and Rob Fergus.
\newblock Indoor segmentation and support inference from rgbd images.
\newblock In {\em Computer Vision -- ECCV 2012}, pages 746--760, 2012.

\bibitem{Song_2015_CVPR}
Shuran Song, Samuel~P. Lichtenberg, and Jianxiong Xiao.
\newblock Sun rgb-d: A rgb-d scene understanding benchmark suite.
\newblock In {\em Proceedings of the IEEE Conference on Computer Vision and
  Pattern Recognition (CVPR)}, 2015.

\bibitem{Wang_2022_CVPR}
Haowen Wang, Mingyuan Wang, Zhengping Che, Zhiyuan Xu, Xiuquan Qiao, Mengshi
  Qi, Feifei Feng, and Jian Tang.
\newblock Rgb-depth fusion gan for indoor depth completion.
\newblock In {\em Proceedings of the IEEE/CVF Conference on Computer Vision and
  Pattern Recognition (CVPR)}, pages 6209--6218, 2022.

\bibitem{10.1007/978-3-031-19812-0_13}
Zhiqiang Yan, Kun Wang, Xiang Li, Zhenyu Zhang, Jun Li, and Jian Yang.
\newblock Rignet: Repetitive image guided network for depth completion.
\newblock In {\em Computer Vision -- ECCV 2022}, pages 214--230, 2022.

\bibitem{Yu_2019_ICCV}
Jiahui Yu, Zhe Lin, Jimei Yang, Xiaohui Shen, Xin Lu, and Thomas~S. Huang.
\newblock Free-form image inpainting with gated convolution.
\newblock In {\em Proceedings of the IEEE/CVF International Conference on
  Computer Vision (ICCV)}, 2019.

\bibitem{Zhang_2018_CVPR}
Yinda Zhang and Thomas Funkhouser.
\newblock Deep depth completion of a single rgb-d image.
\newblock In {\em Proceedings of the IEEE Conference on Computer Vision and
  Pattern Recognition (CVPR)}, 2018.

\bibitem{Zhao_2017_CVPR}
Hengshuang Zhao, Jianping Shi, Xiaojuan Qi, Xiaogang Wang, and Jiaya Jia.
\newblock Pyramid scene parsing network.
\newblock In {\em Proceedings of the IEEE Conference on Computer Vision and
  Pattern Recognition (CVPR)}, 2017.

\bibitem{Zoph_2018_CVPR}
Barret Zoph, Vijay Vasudevan, Jonathon Shlens, and Quoc~V. Le.
\newblock Learning transferable architectures for scalable image recognition.
\newblock In {\em Proceedings of the IEEE Conference on Computer Vision and
  Pattern Recognition (CVPR)}, 2018.

\end{thebibliography}
}

\end{document}